%% file: constrained_bo_arxiv.tex
\documentclass[twoside]{article}
\usepackage{ftnright}

\usepackage[accepted]{aistats2021}

\usepackage{microtype}
\usepackage{graphicx}
\usepackage{booktabs} 
\usepackage{natbib}

\usepackage[hidelinks]{hyperref}
\usepackage{placeins}
\usepackage{todonotes}
\usepackage[group-separator={,}]{siunitx}
\usepackage{xspace}
\usepackage{amsfonts}
\usepackage{amsmath}
\usepackage{subcaption}
\usepackage{multirow}
\usepackage{amssymb,amsmath,amsthm}
\newtheorem{theorem}{Theorem}

\input{macros.tex}

\newcommand{\customfootnotetext}[2]{{
  \renewcommand{\thefootnote}{#1}
  \footnotetext[0]{#2}}}

\begin{document}

\twocolumn[
    \aistatstitle{Scalable Constrained Bayesian Optimization}
    \aistatsauthor{David Eriksson$^1$ \And Matthias Poloczek$^1$}
    \aistatsaddress{Facebook\\\texttt{deriksson@fb.com} \And Amazon\\\texttt{matpol@amazon.com}}
]

\begin{abstract}
\input{abstract.tex}

\end{abstract}

\customfootnotetext{$1$}{This work was conducted while the authors were affiliated with Uber AI.}

\section{Introduction}
\label{sec:introduction}
\input{introduction.tex}

\section{The Model}
\label{sec:model}
\input{model.tex}

\section{Scalable Constrained Bayesian Optimization (\ALG)}
\label{sec:methods}
\input{methods.tex}

\section{Experimental Evaluation}
\label{sec:experiments}
\input{experiments.tex}

\section{Conclusions}
\label{sec:conclusions}
\input{conclusions.tex}

\bibliographystyle{abbrvnat}
\bibliography{references}

\newpage
\appendix
\section*{Supplementary material}
\label{sec:supp}
\input{supplementary.tex}

\end{document}

%% file: macros.tex
\graphicspath{{./figures/}}

\newcommand{\ALG}{\texttt{SCBO}\xspace}
\newcommand{\CEI}{\texttt{cEI}\xspace}
\newcommand{\CMA}{\texttt{CMA-ES}\xspace}
\newcommand{\COBYLA}{\texttt{COBYLA}\xspace}
\newcommand{\RS}{\texttt{RS}\xspace}
\newcommand{\PESC}{\texttt{PESC}\xspace}
\newcommand{\SLACK}{\texttt{SLACK}\xspace}
\newcommand{\TURBO}{\texttt{TURBO}\xspace}

\newcommand{\RANDOM}{\texttt{RS}\xspace}
\newcommand{\TS}{\texttt{TS}\xspace}

\newcommand{\R}{\mathbb{R}}

\DeclareMathOperator*{\argmin}{argmin}

\newcommand{\len}{L}

%% file: abstract.tex
The global optimization of a high-di\-men\-sional black-box function under black-box constraints is a pervasive task in machine learning, control, and engineering.
These problems are challenging since the feasible set is typically non-convex and hard to find, in addition to the curses of dimensionality and the heterogeneity of the underlying functions.
In particular, these characteristics dramatically impact the performance of Bayesian optimization methods, that otherwise have become the de facto standard for sample-efficient optimization in unconstrained settings, leaving practitioners with evolutionary strategies or heuristics.
We propose the scalable constrained Bayesian optimization (\ALG) algorithm that overcomes the above challenges and pushes the applicability of Bayesian optimization far beyond the state-of-the-art.
A comprehensive experimental evaluation demonstrates that \ALG achieves excellent results on a variety of benchmarks.
To this end, we propose two new control problems that we expect to be of independent value for the scientific community.

%% file: introduction.tex
The global optimization of a black-box objective function under black-box constraints has many applications in machine learning, engineering, and the natural sciences.
Examples include fine-tuning the efficiency of a computing platform while preserving the quality of service; maximizing the power conversion efficiency of a solar cell material under stability and reliability requirements; optimizing the control policy of a robot under performance and safety constraints; tuning the performance of an aerospace design averaged over multiple scenarios while ensuring a satisfactory performance on each individual scenario (multi-point optimization).
Moreover, a popular approach for multi-objective optimization tasks is to to reformulate them as constrained problems.
Here the functions that comprise the objective and the constraints are often given as black-boxes, i.e., upon their evaluation we receive an observation of the respective function, possibly with noise but without derivative information.
All of the above examples have in common that their dimensionality, that is, the number of tunable parameters, is large: it is usually up to several dozens, which poses a substantial challenge for current methods in derivative-free optimization.

High dimensionality makes black-box functions hard to optimize due to the curses of dimensionality~\citep{powell2019unified}, even in the absence of constraints.
Moreover, these functions are often heterogeneous which poses a problem for surrogate-based optimizers.
Black-box constraints make the task considerably harder since the set of feasible points is typically non-convex and hard to find, e.g., for control problems.

The main contributions of this work are as follows:
\begin{enumerate}
  \item We propose the \emph{scalable constrained Bayesian optimization} algorithm (\ALG), the first scalable algorithm for the optimization of high-dimensional expensive functions under expensive constraints. \ALG is also the first algorithm to support large batches for constrained problems with native support for asynchronous observations.

  \item A comprehensive evaluation shows that \ALG outperforms previous state-of-the-art methods by far on high-dimensional constrained problems. Moreover, \ALG at least matches and often beats the best performer on low-dimensional instances.

  \item We introduce two new high-dimensional constrained test problems that will be of independent interest given the novelty and anticipated impact of large-scale constrained Bayesian optimization.
\end{enumerate}

\subsection{Related Work}
\label{section_related_work}
Bayesian optimization (BO) has recently gained enormous popularity for the global optimization of expensive black-box functions, see~\citep{frazier2018tutorial,shahriari2016taking} for an overview.
While the vast majority of work focuses on unconstrained problems, aside from box constraints that describe the search space, a handful of articles consider the presence of black-box constraints.
The seminal work of~\citet{schonlau1998global} extends the expected improvement criterion (\texttt{EI}) to constraints by multiplying the expected improvement at some point~$x$ over the best \emph{feasible} point with the probability that~$x$ itself is feasible, leveraging the independence between the objective function and the constraints.
Later this \CEI algorithm was rediscovered by~\citep{cei,gelbart2014bayesian} and studied in a variety of settings, e.g., see~\citep{sobester2014engineering,forrester2008engineering,parr2012infill,parr2012enhancing} and the references therein.
\citet{letham2019constrained} extended the approach to noisy observations using quasi Monte Carlo integration and were the first to consider batch acquisition under constraints.
Note that for noise-free observations, as for the benchmarks that we study, their approach reduces to the original~\CEI.

\citet{pesc} extended predictive entropy search~\citep{hernandez2014predictive} to constraints and detailed how to make the sophisticated approximation of the entropy reduction computationally tractable in practice.
Their \PESC algorithm usually achieves great results and is widely considered the state-of-the-art for constrained BO despite its rather large computational costs.
\citet{picheny2014stepwise} considered the volume of the admissible excursion set under the best known feasible point as a measure for the uncertainty over the location of the optimizer.
His algorithm iteratively samples a point that yields a maximum approximate reduction in volume.

By lifting constraints into the objective via the Lagrangian relaxation, \citet{gramacy2016modeling} took a different approach.  Note that it results in a series of unconstrained optimization problems that are solved by vanilla BO.
\SLACK of \citet{picheny2016bayesian} refined this idea by introducing slack variables and showed that this augmented Lagrangian achieves a better performance for \emph{equality} constraints.
Very recently, \citet{ariafar2019admmbo} used the \texttt{ADMM} algorithm to solve an augmented Lagrangian relaxation.
All these algorithms use the~\texttt{EI} criterion.

\pagebreak

Traditionally, BO, with or without constraints, has been limited to problems with a small number of decision variables, usually at most~15, and a budget of no more than a couple of hundred samples.
Recent work has started exploring scalable BO for budgets with tens of thousands of samples.
\citet{hernandez2017parallel} extended Thompson sampling~\citep{thompson1933likelihood} to large batch sizes and used a Bayesian neural network for the surrogate to maintain scalability (see also~\citep{kandasamy2018parallelised}).
\citet{wang2017batched} proposed the \texttt{EBO} algorithm that partitions the search space to achieve scalability.
\citet{eriksson2019} abandoned a global surrogate and instead maintained several local models that move towards better solutions.
Their \TURBO algorithm applies a bandit approach to allocate samples efficiently between these local searches.
Independently, \citet{mathesen2020stochastic} also proposed to combine trust region modeling with Bayesian optimization with an~\texttt{EI}-based acquisition criterion to balance global and local optimization.
BO has been investigated for high-dimensional settings with small sampling budgets, albeit without constraints, e.g., see~\citep{wang2016bayesian,binois2015warped,binois2020choice,eriksson2018scaling,mutny2018efficient,oh2018bock,rolland2018high,HeSBO19}.
\citet{Bouhlel2018} combine a dimensionality reduction via partial least squares with kriging-based \texttt{EI} to solve a~$50$-dimensional reduced version of the constrained MOPTA problem. The authors point out that their approach cannot handle the full MOPTA problem studied in Sect.~\ref{sec:experiments}.

The constrained optimization of black-box functions has also been studied in the field of evolutionary strategies and in operations research.
\CMA is one of the most powerful and versatile evolutionary strategies.
It uses a covariance adaptation strategy to learn a second-order model of the objective function.
\CMA handles constraints by the 'death penalty' that sets the fitness value of infeasible solutions to zero~\citep{kramer2010review,arnold2012}.
\COBYLA~\citep{Powell1994} and \texttt{BOBYQA}~\citep{powell2007view} maintain a local trust region and thus perform a local search.
In our experience, this strategy scales well to high dimensions, with \COBYLA having an edge due to its support for non-linear constraints.
We will compare to \CEI that we extended to high-dimensional domains, \PESC, \SLACK, \CMA, and \COBYLA and thus have a  representative selection of above lines of work.
Data-dependent transformations of the black-box functions were studied in~\citep{snelson2004warped,wilson2010copula,snoek2014input,salinas2019copula}.

\noindent\textbf{Structure of the article.} The remainder of the article is structured as follows.
In the next section we define the problem formally.
The \ALG algorithm is presented in Sect.~\ref{sec:methods} and compared to a representative selection of methods in Sect.~\ref{sec:experiments}.
Sect.~\ref{sec:conclusions} summarizes the conclusions and discusses ideas for future work.

%% file: model.tex
The goal is to find an optimizer
\begin{equation}
    \label{eq:gop}
    \argmin_{x \in \Omega} f(x) \,\,\, \text{s.t.} \,\,\, c_1(x) \leq 0, \ldots, c_m(x) \leq 0
\end{equation}
where $f{:}\, \Omega \to \R$ and $c_\ell {:}\, \Omega \to \R$ for~$1\,{\leq}\, \ell \,{\leq}\, m$ are black-box functions defined over a compact set~$\Omega \subset \R^d$.
The term black-box function means that we may query any~$x \in \Omega$ to observe the values under the objective function~$f$ and all constraints, possibly with noise, but no derivative information.
Specifically, we suppose that we observe an i.i.d.\ $(m+1)$-dimensional vector with the~$\ell$-th entry given by~$y_0(x) \sim {\cal N}(f(x),\lambda_0(x))$ and~$y_\ell(x) \sim {\cal N}(c_\ell(x),\lambda_\ell(x))$ for~$1\,{\leq}\, \ell \,{\leq}\, m$.
Here the~$\lambda$'s give the variance of the observational noise and are supposed to be known.
In practice, we estimate the~$\lambda$'s along with the hyperparameters of the surrogate model.
Note that we may rescale the search space~$\Omega$ w.l.o.g.\ to the unit hypercube $[0, 1]^d$.
If all functions are observed without noise, then our goal is to find a feasible point with minimum value under the objective function.
For noisy functions, we wish to find a point with best expected objective value under all points that are feasible with probability at least~$1{-}\delta$, where~$\delta$ is set based on the context, e.g., the degree of risk aversion (cp.\ \citet{pesc}).

%% file: methods.tex
We propose the \emph{Scalable Constrained Bayesian Optimization} (\ALG) algorithm.
\ALG follows the paradigm of the generic BO algorithm~\citep{frazier2018tutorial,shahriari2016taking} and proceeds in rounds.
In each round, \ALG selects a batch of~$q$ points in~$\Omega$ that are then evaluated in parallel.
Note that~\ALG is easily extended to asynchronous batch evaluations.

\ALG employs the \emph{trust region} approach introduced by~\citet{eriksson2019}
that confines samples locally.
This addresses common problems of Bayesian optimization in high-dimensional settings, where popular acquisition functions spread out samples due to the inherently large uncertainty and thus fail to zoom in on promising solutions.
Moreover, for the popular Mat\'ern kernels, the covariance under the prior is essentially zero for two points if they differ substantially in one coordinate only.
The use of trust regions results in more exploitation and often a better fit for the local surrogate.
\ALG maintains the invariant that the trust region is centered at a point of maximum utility.
Thus, the trust region is moved through the domain~$\Omega$ as better points are discovered.

The generalization to black-box constraints poses additional fundamental problems that were not considered by~\citet{eriksson2019}.
For many problems it is hard to even find a feasible solution, since the feasible set is typically non-convex.
An investigation in Sect.~\ref{sec:experiments} demonstrates the difficulty of this task.
Moreover, black-box functions often vary drastically in their characteristics across~$\Omega$.
We will provide examples where some constraints exhibit a huge variability whereas others are smooth.
\ALG applies tailored \emph{transformations} that account for the specific roles of the objective and the constraints.

\noindent\textbf{Extending Thompson sampling to constrained optimization.}
\ALG extends Thompson sampling (\TS) to black-box constraints, and is to the best of our knowledge the first to do so.
\TS scales to large batches at low computational cost and is at least as effective as \texttt{EI}, as we demonstrate below.
To select a point for the next batch, \ALG samples~$r$ candidate points in~$\Omega$ (see the supplement for details).

Let~$x_1,\ldots, x_r$ be the sampled candidate points.
Then \ALG samples a \emph{realization}~$(\hat{f}(x_i), \hat{c}_1(x_i), \ldots,\hat{c}_m(x_i))^T$ for all~$x_i$ with~$1\leq i\leq r$ from the respective posterior distributions on the functions~$f,c_1,\ldots,c_m$.
Let~$\hat{F} = \{x_i \mid{} \hat{c}_\ell(x_i) \leq 0 \text{ for } 1\leq \ell \leq m\}$ be the set of points whose realizations are feasible.
If~$\hat{F} \neq \emptyset$ holds, \ALG selects an~$\argmin_{x \in \hat{F}} \hat{f}(x)$.
Otherwise \ALG selects a point of \emph{minimum total violation}~$\sum_{\ell=1}^m \max\{\hat{c}_\ell(x), 0\}$, breaking ties via the sampled objective value.
While we found that this natural selection criterion is able to identify a feasible point quickly for smooth constraints, we observed that it struggles when functions vary significantly in their magnitudes.

\begin{figure*}[!t]
    \centering
    \includegraphics[width=0.84\textwidth]{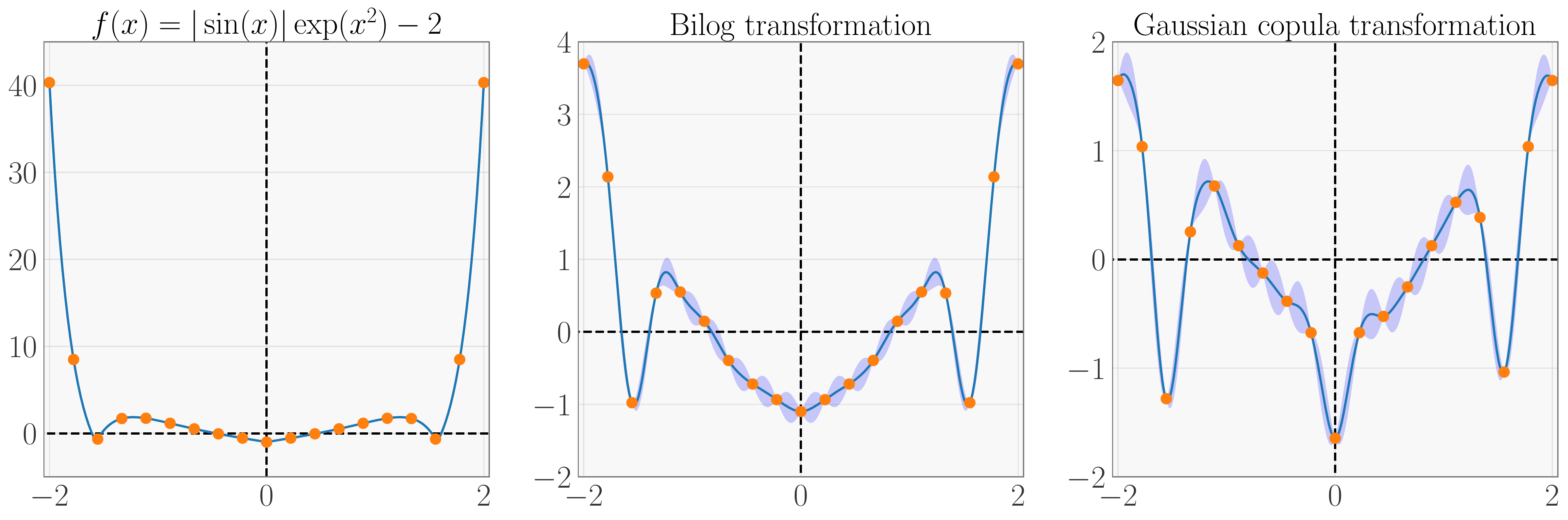}
    \caption{
        \textbf{(Left)} The original function where the distance to the origin varies considerably for the observations.
        If this was a constraint, the feasible region, denoted by the change of the sign, would be hard to detect.
        If it was the objective function, we would struggle to identify the minima, since the observations in the center differ only slightly and are considerably smaller in absolute value than the observations on the boundary.
        \textbf{(Middle)} The bilog transformation stretches out observations around zero, thereby making it easier to detect feasible areas.
        Note that a GP has been fitted to the observations given by the orange points in the middle and the right plot.
        The blue line depicts the posterior mean and the shaded area gives the posterior uncertainty of the GP.
        \textbf{(Right)} The copula transformation magnifies values that are at the ends of the observed spectrum, which facilitates the task of finding optima.
        Note that these transformations are advantageous over a naive standardization of each function as the latter is insensitive to the areas of interest.
    }
    \label{fig:transforms}
    \raggedbottom
\end{figure*}

\noindent\textbf{Transformations of objective and constraints.}
The key observations are that for the objective function we are particularly interested in the locations of possible optima, whereas for constraints we are interested in identifying feasible areas, i.e., where the constraint function values become negative.
Thus, we apply transformations that emphasize these areas particularly; see also Fig.~\ref{fig:transforms}.
To the objective function, we apply a \emph{Gaussian copula} \citep[e.g., see][]{wilson2010copula}.
The Gaussian copula first maps all observations under the objective to quantiles using the empirical CDF.
Then it maps the quantiles through an inverse Gaussian CDF.
Note that this procedure magnifies differences between values that are at the end of the observed range, i.e., minima or maxima.
It affects the observed values but not their location.
Finally, we apply Gaussian process regression to the mapped observations, as usual.
For the constraints we employ the \texttt{bilog} transformation: $\mathrm{bilog}(y) = \mathrm{sgn}(y) \ln{(1 + |y|)}$ for a scalar observation~$y$.
It magnifies the range around zero to emphasize the change of sign that is decisive for feasibility.
Moreover, it dampens large values.

\noindent\textbf{Maintaining the trust region.}
The trust region is initialized as a hypercube with side length~$L = L_\text{init}$.
We count for each trust region the number of \emph{successes}~$n_s$ and \emph{failures}~$n_f$ since it was resized last.
First suppose that all functions are observed without noise.
Then a success occurs when \ALG observes a better point; by construction, this point must be inside the trust region.
A failure happens when no point in the batch is better than the current center of the trust region.
The center~$C$ of the trust region is chosen as follows.
We select the best feasible point for~$C$ if any. Otherwise we pick a point with minimum total violation, again breaking ties via the objective.
Note that we use (transformed) \emph{observations} from the black-box functions, not \emph{realizations} from the posterior.
Thus, the center is moved to a new point whenever a success occurs.
The trust region is resized as follows: if~$n_s = \tau_s$ then the side length is set to~$L = \min\{2L, L_{\max}\}$ and we reset $n_s = 0$.
If~$n_f = \tau_f$, then we set~$L = L/2$ and $n_f=0$.
If the side length drops below a set threshold~$L_{\min}$, then we initialize a new trust region.
For noisy functions we follow the same rules, and use the posterior mean of GP model instead of the observed value.
Note that the procedure for maintaining the trust regions follows~\citep{eriksson2019} and is described here for completeness.
In the next section we demonstrate that \ALG achieves excellent performance across all benchmarks.

\subsection{Summary of the \ALG Algorithm}
We summarize the \ALG algorithm.
\begin{enumerate}
    \item Evaluate an initial set of points and initialize the trust region at a point of maximum utility.
    \item Until the budget for samples is exhausted:
    \begin{enumerate}
        \item Fit GP models to the transformed observations.
        %
        \item Generate~$r$ candidate points~$x_1,\ldots,x_r \in \Omega$ in the trust region.
        \item For each of the~$q$ points of the next batch we sample a realization \\
        $\{(\hat{f}(x_i), \hat{c}_1(x_i), \ldots,\hat{c}_m(x_i))^T \mid{} 1{\leq} i {\leq} r \}$
        from the posterior over each candidate and add a point of maximum utility to the batch.
        \item Evaluate the objective and constraints at the~$q$ new points.
        \item Adapt the trust region by moving the center as described above.
        Update the counters~$n_s$, $n_f$, and size~$L$.
        If~$L < L_{\min}$, initialize a new trust region.
    \end{enumerate}
    \item Recommend an optimal feasible point (if any).
\end{enumerate}

For noisy functions, we recommend a point of minimum posterior mean under all points that are feasible with probability at least~$1-\delta$ (if any).
Note that~\ALG is \emph{consistent} and hence will converge to a global optimum as the number of samples tends to infinity.
The proof was deferred to the supplement due to space constraints.

%% file: experiments.tex
We compare \ALG to the state-of-the-art: \PESC~\citep{pesc} in \texttt{Spearmint},
\CEI~\citep{schonlau1998global,cei},
\SLACK~\citep{picheny2016bayesian} in \texttt{laGP},
the implementation of~\cite{jones2014scipy} for \COBYLA~\citep{Powell1994},
\CMA~\citep{hansen2006cma} in \texttt{pycma}, and random search (\RANDOM).
Please see Sect.~\ref{section_related_work} for a discussion of these methods.

\noindent\textbf{The Benchmarks.} We evaluate the algorithms on a comprehensive selection of benchmark problems.
First, we consider four low-dimensional problems in Sect.~\ref{section_physics_problems}: a $3$D tension-compression string problem with four constraints, a $4$D pressure vessel design with with four constraints, a $4$D welded beam design problem with five constraints, and a $7$D speed reducer problem with eleven constraints.
Next we consider the $10$D Ackley problem with two constraints in Sect.~\ref{section_ackley} that is particularly interesting because of its small feasible region.
Then we study four large-scale problems: the $30$D Keane bump function with two constraints in Sect.~\ref{section_keane}, a $12$D robust multi-point optimization problem with a varying number of constraints in Sect.~\ref{section_lunar}, a $60$D trajectory planning problem with $15$ constraints in Sect.~\ref{section_rover}, and a $124$D vehicle design problem with $68$ constraints in Sect.~\ref{section_mopta}.
\PESC and \SLACK do not scale to large-scale high-dimensional problems and large batch sizes and are therefore omitted for these problems.
Note that all benchmarks have multi-modal objective functions and are observed without noise.
We perform $30$ replications for each experiment.

To compare feasible and infeasible solutions, we adopt the rationale of \citet{pesc} that any feasible solution is preferable over an infeasible one and thus assign a default value to infeasible solutions that is set to the largest found objective value for the respective benchmark.
Performance plots show the mean with one standard error.
All methods start with an initial set of points given by a Latin hypercube design (LHD).
\CMA and \COBYLA are initialized from the best point in this design.
Recall that \ALG applies transformations to the functions.
In the supplement we investigate the performances of the baselines under these transformations and show that \ALG performs best.

\subsection{Physics Test Problems}
\label{section_physics_problems}
\begin{figure*}[!ht]
        \centering
        \includegraphics[width=0.85\textwidth]{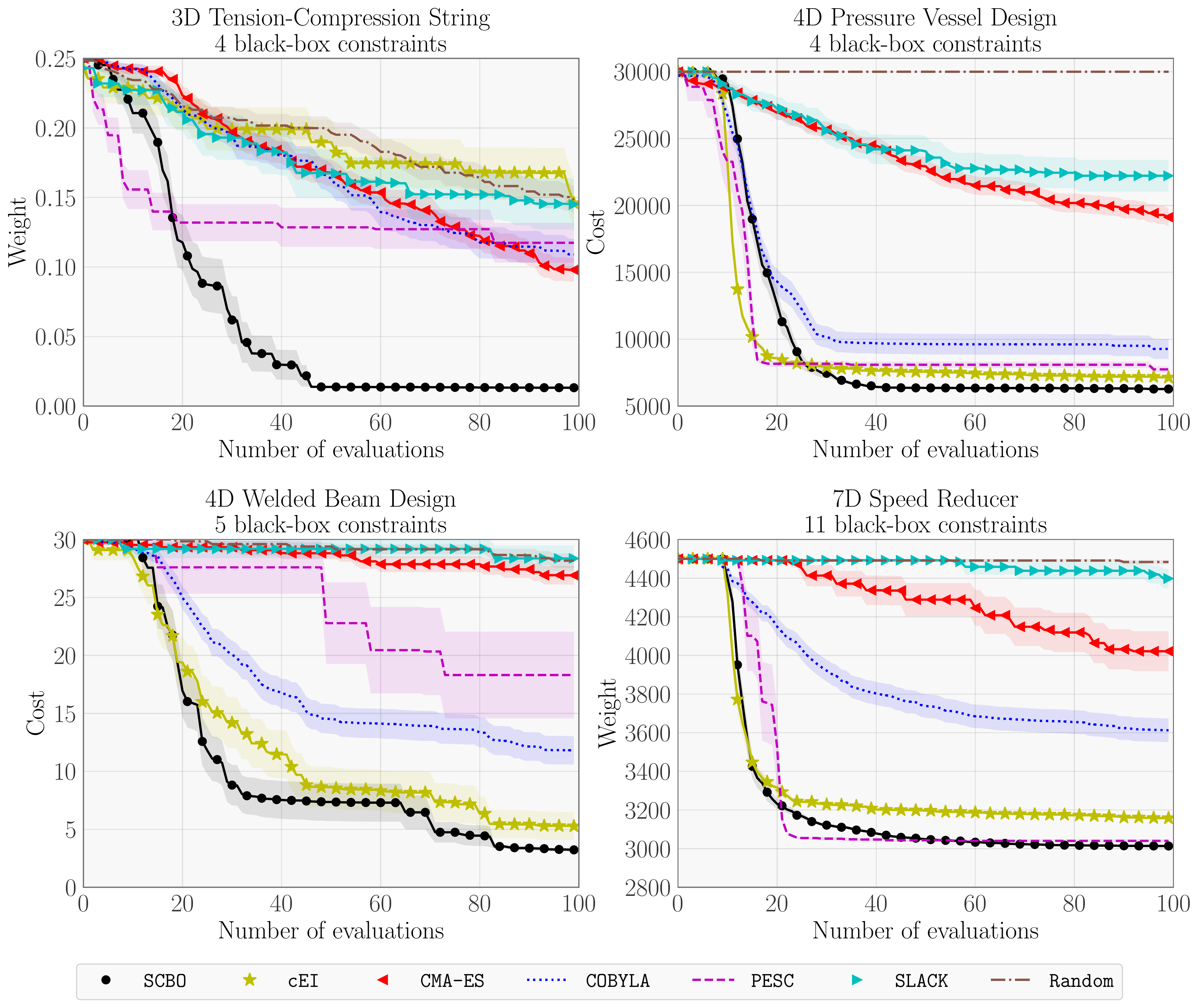}
        \caption{
            \textbf{(Upper left)} \ALG outperforms the other methods on the Tension-compression string problem.
            \textbf{(Upper right)} \ALG finds the best solutions on the pressure vessel design problem, followed by \CEI, \PESC, and \COBYLA.
            \textbf{(Lower left)} \ALG performs best on the welded beam design problem.
            \textbf{(Lower right)} \ALG and \PESC perform the best on the speed reducer problem.
        }
        \label{fig:physics}
        \raggedbottom
\end{figure*}

We evaluate the algorithms on a variety of physics problems.
We use a budget of~$100$ evaluations, batch size~$q=1$, and~$10$ initial points.
Fig.~\ref{fig:physics} summarizes the results for the four test problems.
\ALG outperforms all baselines on the $3$D tension-compression string problem~\citep{hedar2006derivative}:
it found feasible solutions in all runs and consistently obtained excellent solutions.
\PESC and \CEI are not competitive. Their performance is only slightly better than \RANDOM search on this problem.
For the $4$D pressure vessel design problem~\citep{coello2002constraint}, \ALG obtains the best solutions followed by \CEI, \PESC, and \COBYLA.
\ALG also performs best for the $4$D welded beam design problem~\citep{hedar2006derivative}, followed by~\CEI.
\ALG and \PESC obtain excellent results for the $7$D speed reducer design problem~\citep{lemonge2010constrained}.

\begin{figure*}[!t]
    \centering
    \includegraphics[width=0.82\textwidth]{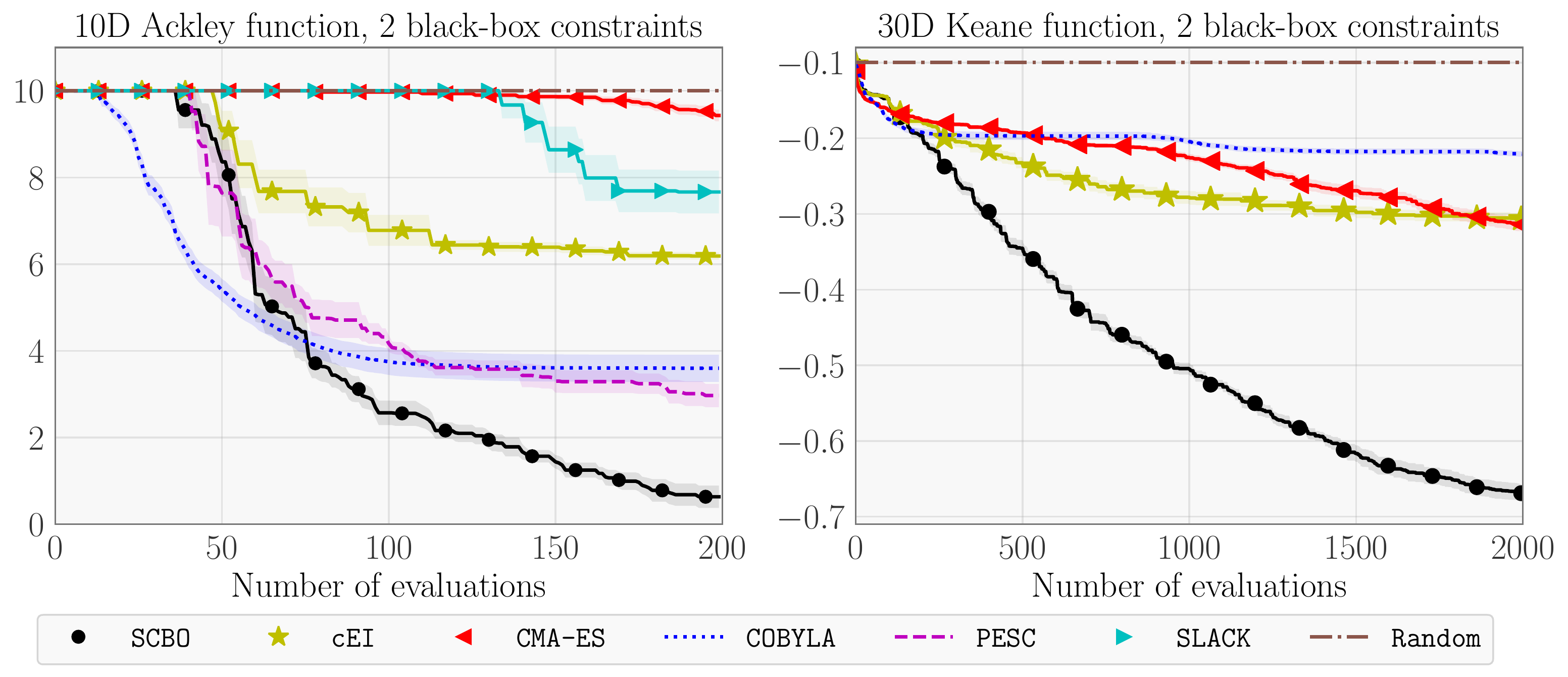}
    \caption{\textbf{(Left)} 10D Ackley function with two constraints. \ALG consistently finds solutions close to the global optimum.
    \textbf{(Right)} 30D Keane function with two constraints.
    \ALG clearly outperforms the other methods from the start.}
    \label{fig:synthetic}
    \raggedbottom
\end{figure*}

\subsection{The 10D Ackley Function}
\label{section_ackley}
We study the performance on the $10$D Ackley function on the domain $[-5, 10]^{10}$ with the constraints $c_1(x) = \sum_{i=1}^{10} x_i \leq 0$ and $c_2(x) = \|x\|_2 - 5 \leq 0$.
The Ackley function has a global optimum with value zero at the origin.
This is a challenging problem where the probability of randomly selecting a feasible point is only~$2.2 \cdot 10^{-5}$.
We use a budget of~$200$ function evaluations, batch size~$q=1$, and $10$ initial points.
Fig.~\ref{fig:synthetic} shows that \COBYLA initially makes good progress but is eventually outperformed by \ALG which achieves the best performance.
\PESC performs well, but is computationally costly: a run with \PESC took $3$ hours, while the other methods ran in minutes.

\subsection{The 30D Keane Bump Function}
\label{section_keane}
The Keane bump function is a common test function for constrained global optimization~\citep{keane1994experiences}.
This function has two constraints.
We consider $d=30$ and batch size~$50$ for \ALG, \CEI, and \CMA.
Each method uses~$100$ initial points.
\COBYLA does not support batching samples and thus samples sequentially, which is an advantage as it can leverage more data for acquisition.
However, Fig.~\ref{fig:synthetic} shows that nonetheless \COBYLA is not competitive.
We see that \ALG clearly outperforms the other algorithms for this challenging high-dimensional benchmark.
As stated above, we cannot compare to \PESC and \SLACK on this large-scale benchmark due to their computational overhead.

\subsection{Robust Multi-point Optimization}
\label{section_lunar}
Multi-point optimization is an important task in aerospace engineering~\citep{liem2014multimission,liem2017expected,quim2018}.
Here, a design is optimized over a collection of flight conditions.
Multi-point optimization produces designs with better practical performance by addressing the issue that tuning a design for a single
scenario often leads to designs with poor off-scenario performance~\citep{jameson1990automatic,cliff2001single}.
In this section we propose a \emph{robust multi-point optimization problem}.
The goal is to optimize the performance of the design~$x$ averaged over~$m$ scenarios (potentially weighted by importance), subject to individual constraints that assert an acceptable performance for each scenario.
Our problem is derived from the lunar lander problem, where the goal is to find a $12$D controller that maximizes the reward averaged over~$m$ terrains.
We extend this problem by adding~$m$ constraints that assert that no individual reward is below~$200$, which guarantees that the lunar lands successfully.
Without these constraints, the algorithms often produce policies that occasionally crash the lander.
We evaluate the algorithms with~$1000$ samples, batch size~$q=50$, and~$50$ initial points for three experiments that differ in the number of constraints:~$m=10$, ~$m=30$, and~$m=50$.
Tab.~\ref{tab:lunar} summarizes the results.

\begin{table}[!ht]
    \centering
    \resizebox{0.44\textwidth}{!}{
        \begin{tabular}{|c|c|ccccc|}
            \hline
            $m$ &                          & \ALG           & \CEI  & \CMA  & \COBYLA & \RANDOM \\
            \hline
            \multirow{5}{*}{10} & Best     & 321.3 & 322.2 & 310.3 & 315.0 & \text{NA} \\
                                & Worst    & 302.8 & 250.9 & 266.4 & 293.0 & \text{NA} \\
                                & Median   & 318.0 & 318.5 & 291.9 & 299.7 & \text{NA} \\
                                & Feasible & 28/30 & 27/30 & 21/30 & 2/30  & 0/30 \\
            \hline
            \multirow{5}{*}{30} & Best     & 316.2 & 311.2 & 293.2 & 312.7 & \text{NA} \\
                                & Worst    & 295.2 & 267.2 & 270.3 & 294.2 & \text{NA} \\
                                & Median   & 311.8 & 288.4 & 283.7 & 295.2 & \text{NA} \\
                                & Feasible & 26/30  & 10/30 & 11/30 & 5/30  & 0/30 \\
            \hline
            \multirow{5}{*}{50} & Best     & 309.7 & 295.4 & 295.7 & 300.0 & \text{NA} \\
                                & Worst    & 276.2 & 262.7 & 256.3 & 276.5 & \text{NA} \\
                                & Median   & 306.0 & 269.9 & 274.5 & 285.1 & \text{NA} \\
                                & Feasible & 27/30  & 8/30  & 11/30  & 3/30  & 0/30 \\
            \hline
        \end{tabular}
    }
    \caption{Results for the 12D multi-point optimization problem.
    %
    %
    We observe that \ALG finds the best robust policies over thirty runs and scales best to larger numbers of constraints.}
    \label{tab:lunar}
\end{table}

We see that \ALG found a feasible controller for 28/30 runs when~$m=10$.
\CEI and \CMA found feasible points in 27/30 and 21/30 runs respectively.
\COBYLA and \RANDOM struggled visibly.
We note that \ALG clearly outperformed the other methods for the more constrained settings with~$m=30$ and~$m=50$.
Note that the best reward found by each algorithm clearly decreases when~$m$ increases.
The problem becomes harder when we add more scenarios, since feasible solutions for $m=10$ may not satisfy all constraints for~$m=30$ or~$m=50$.

\subsection{60D Rover Trajectory Planning}
\label{section_rover}
We study a $60$D route planning problem adapted from~\citep{wang2017batched}.
The task is to position $30$ waypoints that lead a rover on a path of minimum cost from its starting position to its destination, while avoiding collisions with obstacles.
We propose a constraint-based extension with~$m=15$ constraints that are met if and only if the rover does not collide with any associated impassable obstacles.
The exact formulation of these constraints is given in the supplementary material.
Fig.~\ref{fig:rover_mopta} (middle) illustrates the setup and the best trajectory found by \ALG.
\begin{figure*}[!ht]
    \centering
    \includegraphics[width=0.85\textwidth]{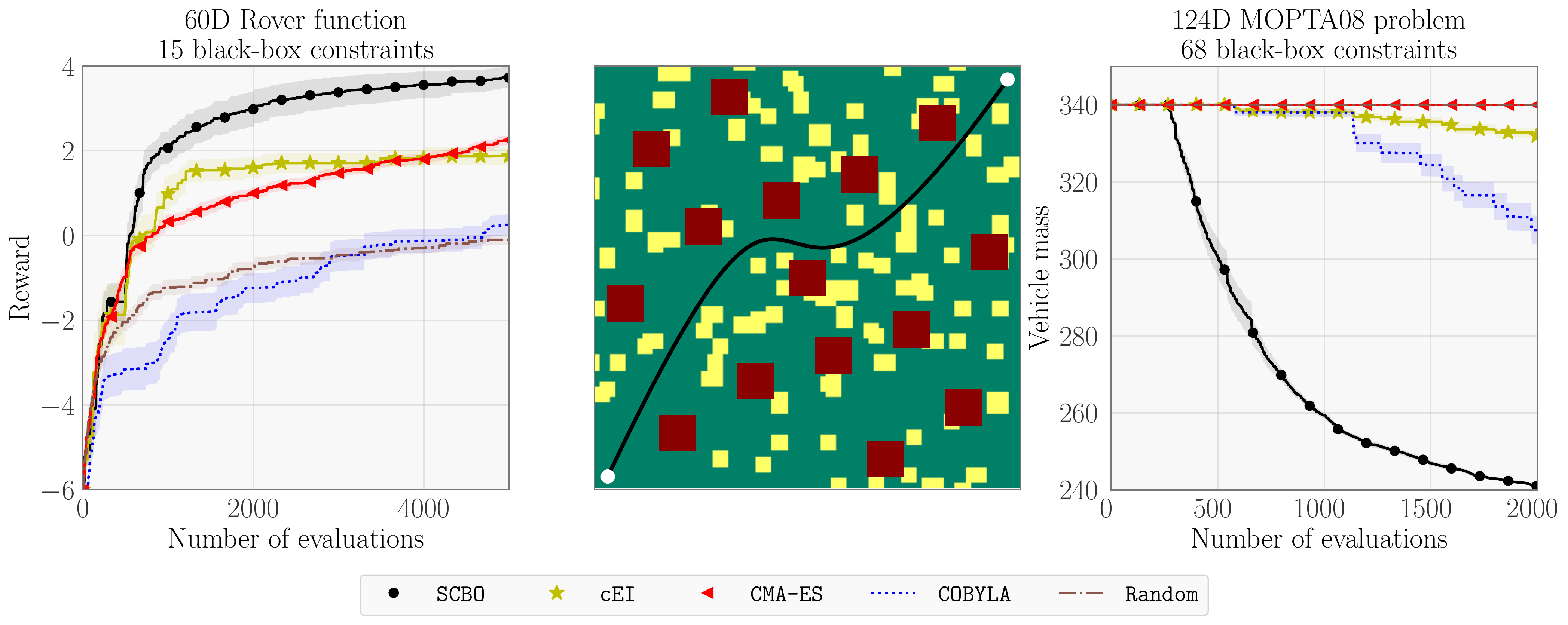}
    \caption{
        \textbf{(Left) 60D trajectory planning with 15 constraints:}~\ALG finds excellent solutions quickly and outperforms the other methods.
        \textbf{(Middle) Illustration of the trajectory planning problem}:
        The black line is the best trajectory found by \ALG with a reward of 4.93.
        The green area can be traversed at no cost.
        Yellow squares denote terrain that inflicts a cost upon traversal.
        Red squares are impassable obstacles.
        \textbf{(Right) 124D Vehicle Design with 68 Constraints}: \ALG finds a feasible point in $30$/$30$ runs and consistently finds good solutions.
        }
    \label{fig:rover_mopta}
    \raggedbottom
\end{figure*}
There are two types of terrain that vary in their cost: the green terrain can be traversed at cost zero and a yellow terrain that inflicts a certain cost.
This problem turns out to be challenging for small sampling budgets.
Thus, we have evaluated \ALG, \CEI, \CMA, \COBYLA, and \RS for a total of~$5000$ evaluations with batch size~$q=100$ and~$100$ initial points.
Fig.~\ref{fig:rover_mopta} (left) summarizes the performances.
We see that \ALG outperforms the other methods by far on this hard benchmark.

\subsection{124D Vehicle Design with 68 Constraints}
\label{section_mopta}
We evaluate the algorithms on a~$124$-di\-men\-sional vehicle design problem MOPTA08~\citep{mopta08}, where the goal is to minimize the mass of a vehicle subject to~$68$ performance constraints.
The~$124$ variables describe gages, materials, and shapes.
We ran all experiments with a budget of~$2000$ samples, batch size~$q{=}10$, and~$130$ initial points.
We point out that this benchmark showcases the scalability of the implementation of \ALG that uses GPyTorch \citep{gardner2018gpytorch} and KeOPS \citep{keops} to fit the $69$ GP models in a batch; see the supplement for details.
Fig.~\ref{fig:rover_mopta} (right) shows \ALG, \CEI, \CMA, and \COBYLA over $30$ runs.
\ALG found a feasible solution in all $30$ runs and the best solution found by \ALG had value $236.7$.
\COBYLA found a feasible point in only $11/30$ runs, one which had objective value $238.8$, while \CEI was not able to find a feasible solution with function value below $300$.

\subsection{Ablation studies}
\label{sec:ablation}

\begin{figure*}[!ht]
    \centering
    \includegraphics[width=0.85\textwidth]{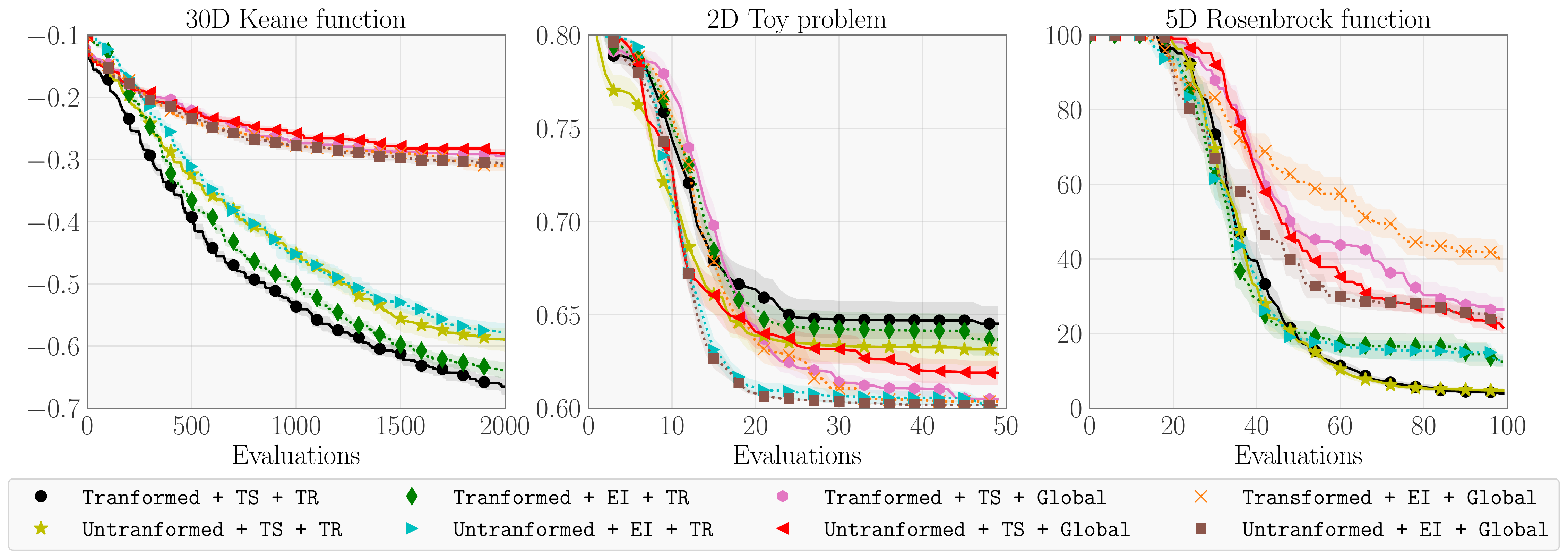}
    \caption{
        We investigate the effects of (i) the transformations, (ii) different acquisition functions (TS/EI), and (iii) the trust region (TR).
        \textbf{(Left)} $30$D Keane function with $2$ black-box constraints.
        \textbf{(Middle)} $2$D Toy problem with $2$ black-box constraints of \citet{pesc}.
        \textbf{(Right)} $5$D Rosenbrock function with $2$ poorly scaled black-box constraints.
    }
    \label{fig:ablation}
    \raggedbottom
\end{figure*}
We investigate how the various components in~\ALG contribute to the overall performance: specifically, how does the application of i) the transformations (Transformed, Untransformed), ii) the acquisition criterion (\TS or \texttt{EI}), and iii) the use of a trust region (TR, Global) affect the performance.
Fig.~\ref{fig:ablation} summarizes the performances of all eight combinations on three benchmarks.
On the left, we see that the use of the trust region is critical for the $30$D Keane function.
Approaches that do not use a trust region struggle, just as in Sect.~\ref{section_mopta}.
Moreover, the transformation provides an additional gain, whereas the choice of the acquisition function has no noticeable effect.
The center plot is for the $2$D toy problem proposed by \citet{pesc} that has a smooth objective function and two easy constraints.
Here, BO without a trust region and \texttt{EI} should shine, and this is indeed the case.
The right plot considers the $5$D Rosenbrock function with two poorly scaled constraints.
Again the trust region is critical for a good performance, as fitting a global surrogate model seems challenging.
Interestingly, \TS achieves significantly better results than \texttt{EI}.

%% file: conclusions.tex
We studied the task of optimizing a black-box objective function under black-box constraints that has numerous applications in machine learning, control, and engineering.
We found that the existing methods struggle in the face of multiple constraints and more than just a few decision variables.
Therefore, we proposed the \emph{Scalable Constrained Bayesian Optimization} (\ALG) algorithm that leverages tailored transformations of the underlying functions together with the trust region approach of \citet{eriksson2019} and Thompson sampling (TS) to scale to high-dimensional spaces and large sampling budgets.

We performed a comprehensive experimental evaluation that compared \ALG to the state-of-the-art from machine learning, operations research, and evolutionary algorithms on a variety of benchmark problems that span control, multi-point optimization, and physics.
We found that \ALG outperforms the state-of-the-art on high-dimensional benchmarks, and matches or beats the performance of the best baseline otherwise.
In the supplement, we also provide an efficient GPU implementation of \ALG based on batch-GPs and a formal proof that \ALG converges to a global optimum.

\pagebreak

For future work, we are interested in applications where the objective and constraints have substantial correlations.
For example, consider the design of an aircraft wing: here the aerodynamic performance (e.g., lift and drag), the structural stability, and the fuel-burn will be related.
If the airfoil’s geometry generates turbulent structures, the drag will increase and the fuel burn will suffer.
The heterogeneity of the involved functions may make the adoption of a multi-output Gaussian process challenging.
We believe that leveraging these correlations may pave an avenue towards solving problems with hundreds of constraints more efficiently.
Constraints also arise naturally for combinatorial black-box functions \citep{baptista2018bayesian,oh2019combinatorial} that have exciting applications in engineering and science.

Moreover, we look forward to inter-disciplinary applications:
\ALG's ability to optimize high-dimensional constrained problems will allow to optimize an airfoil described by a mesh or the parameters controlling a chemical process, e.g., for growing nanotubes or when searching for a solar cell material \citep{herbol2018efficient,ortoll2019bypassing}.

%% file: supplementary.tex
In Sect.~\ref{sec:gps} we describe how we leverage scalable GP regression to run \ALG and \CEI with large sampling budgets.
The value and scalability of our implementation is demonstrated by an experiment in Sect.~\ref{sec:batch_GPs}.
We summarize the hyperparameters of \ALG in Sect.~\ref{sec:scbo_details} and give additional details on how we shrink and expand the trust region.
We prove a consistency result for~\ALG in Sect.~\ref{sec:global_convergence}.
In Sect.~\ref{sec:physics_transformed} we show results for all baselines on the physics test problems where the objective and constraints have been transformed in the same fashion as in~\ALG.
Finally, Sect.~\ref{sec:test_problems} provides details on all benchmarks.

\section{Gaussian process regression}
\label{sec:gps}

As usual, the hyperparameters of the Gaussian process (GP) model are fitted by optimizing the log-marginal likelihood.
The domain is rescaled to \smash{$[0, 1]^d$} and the function values are standardized to have mean zero and variance one before fitting the GP\@.
We use a Mat\'ern-5/2 kernel with ARD and a constant mean function that we optimize using L-BFGS-B.
Following~\citet{snoek2012practical}, a horseshoe prior is placed on the noise variance.
We also learn a signal variance of the kernel.

Scaling BO to large number of evaluations is challenging due to the computational costs of inference.
To compute the posterior distribution for~$n$ observations, we need to solve linear systems with an $n \times n$ kernel matrix.
This is commonly done via a Cholesky decomposition which has a computational complexity of~$\Theta(n^3)$ flops.
When there are $m$ constraints, the cost increases to~$\Theta(m n^3)$ flops and thus may not scale to the large sampling budgets that we consider in this work.
Thus, we leverage the parallelism of modern GPUs that allows to 'batch' several GP models which is provided in the \texttt{GPyTorch} package~\citep{gardner2018gpytorch}.
Relying only on fast matrix vector multiplication (MVM), we can solve linear systems with the kernel matrix using the conjugate gradient (CG) method and approximate the log-determinant via the Lanczos process \cite{dong2017scalable,ubaru2017fast}.
\texttt{GPyTorch} extends this idea to a batch of GPs by computing fast MVMs with a 3D tensor representing a kernel matrix of size $(m+1) \times n \times n$, where~$m{+}1$ is the number of batched GPs.
The MVMs are further sped up by a compiled CUDA kernel constructed via \texttt{KeOPS}~\citep{keops}.
To the best of our knowledge, \ALG is the first Bayesian optimization algorithm to leverage batch GPs and \texttt{KeOPS}.
Note that we also applied these ideas to scale \CEI of~\citet{schonlau1998global} to a large numbers of samples.

\section{Achieving Efficiency via Batch GPs}
\label{sec:batch_GPs}
Next we describe an experiment that demonstrates that the Cholesky decomposition, which is commonly used in GP regression, does not scale to the large sampling budgets that are required for the demanding benchmarks that we study in this work.
We consider training GPs with a different number of training points.
The true function is standardized and we assume observations are subject to  normally distributed noise with mean zero and variance~$0.01$.
We compare the computational cost of the Cholesky decomposition to the efficient batch GP implementation of \texttt{GPyTorch}, both in single precision.
Table~\ref{tab:batchGPs} provides run times for training, predictions, and sampling.
We use~$50$ gradient steps for training. All computations were performed on an NVIDIA RTX 2080 TI.
The two rightmost columns also show the error for the mean and variance predictions.
We see that batch GPs achieve a large speed-up while preserving a high accuracy.
For example, with the batch GP implementation fitting one GP with 1000 points takes 2.33 seconds, while fitting 100 GPs in a batch only takes 11.15 seconds.
Moreover, the Cholesky approach becomes impractical in the large-data regime: training 100 GPs with 8000 points takes 37 minutes, compared to just about 2 minutes for the batch GP implementation!

\begin{table*}[htb]
    \centering
    \resizebox{0.98\textwidth}{!}{
        \begin{tabular}{|ll|lll|lll|ll|}
        \hline
        \multicolumn{2}{|c|}{Data size} & \multicolumn{3}{c|}{Cholesky decomposition} & \multicolumn{3}{c|}{Scalable batch GPs} & \multicolumn{2}{c|}{Prediction error} \\
        \#GPs & Training points & Training & Prediction & Sampling & Training & Prediction & Sampling & Mean MAE & Variance MAE\\
        \hline
        1 & 1000 & \SI{0.60}{\second} & \SI{0.06}{\second} & \SI{0.07}{\second} & \SI{2.09}{\second} & \SI{0.04}{\second} & \SI{0.39}{\second} & 1.41e-03 & 4.44e-03 \\
        1 & 2000 & \SI{1.03}{\second} & \SI{0.08}{\second} & \SI{0.09}{\second} & \SI{2.17}{\second} & \SI{0.05}{\second} & \SI{0.39}{\second} & 2.12e-03 & 5.23e-04 \\
        1 & 4000 & \SI{3.67}{\second} & \SI{0.16}{\second} & \SI{0.14}{\second} & \SI{2.26}{\second} & \SI{0.06}{\second} & \SI{0.41}{\second} & 2.22e-04 & 4.19e-06 \\
        1 & 8000 & \SI{22.87}{\second} & \SI{0.49}{\second} & \SI{0.32}{\second} & \SI{4.73}{\second} & \SI{0.13}{\second} & \SI{0.45}{\second} & 3.94e-04 & 6.75e-05 \\
        \hline
        10 & 1000 & \SI{5.98}{\second} & \SI{0.47}{\second} & \SI{0.58}{\second} & \SI{2.49}{\second} & \SI{0.22}{\second} & \SI{1.16}{\second} & 4.74e-02 & 1.00e-01 \\
        10 & 2000 & \SI{10.13}{\second} & \SI{0.76}{\second} & \SI{0.77}{\second} & \SI{3.61}{\second} & \SI{0.31}{\second} & \SI{1.30}{\second} & 5.00e-02 & 6.14e-02 \\
        10 & 4000 & \SI{35.96}{\second} & \SI{1.53}{\second} & \SI{1.26}{\second} & \SI{7.81}{\second} & \SI{0.35}{\second} & \SI{0.95}{\second} & 3.42e-04 & 9.89e-05 \\
        10 & 8000 & \SI{227.30}{\second} & \SI{4.72}{\second} & \SI{2.88}{\second} & \SI{17.73}{\second} & \SI{0.70}{\second} & \SI{1.08}{\second} & 5.24e-05 & 7.07e-06 \\
        \hline
        50 & 1000 & \SI{24.48}{\second} & \SI{2.40}{\second} & \SI{2.96}{\second} & \SI{7.37}{\second} & \SI{0.50}{\second} & \SI{2.77}{\second} & 5.03e-03 & 3.38e-02 \\
        50 & 2000 & \SI{49.02}{\second} & \SI{3.72}{\second} & \SI{3.88}{\second} & \SI{9.65}{\second} & \SI{0.90}{\second} & \SI{3.12}{\second} & 4.06e-03 & 1.64e-03 \\
        50 & 4000 & \SI{184.61}{\second} & \SI{7.90}{\second} & \SI{6.40}{\second} & \SI{21.15}{\second} & \SI{1.75}{\second} & \SI{3.57}{\second} & 1.49e-04 & 2.10e-05 \\
        50 & 8000 & \SI{1134.58}{\second} & \SI{25.37}{\second} & \SI{14.36}{\second} & \SI{66.50}{\second} & \SI{3.72}{\second} & \SI{4.40}{\second} & 5.74e-04 & 1.15e-04 \\
        \hline
        100 & 1000 & \SI{55.94}{\second} & \SI{5.09}{\second} & \SI{6.11}{\second} & \SI{10.41}{\second} & \SI{1.03}{\second} & \SI{6.85}{\second} & 1.85e-03 & 9.22e-04 \\
        100 & 2000 & \SI{108.00}{\second} & \SI{8.42}{\second} & \SI{8.38}{\second} & \SI{18.59}{\second} & \SI{2.27}{\second} & \SI{8.61}{\second} & 2.12e-02 & 2.40e-02 \\
        100 & 4000 & \SI{365.88}{\second} & \SI{16.64}{\second} & \SI{12.54}{\second} & \SI{44.08}{\second} & \SI{3.91}{\second} & \SI{8.42}{\second} & 4.56e-04 & 1.25e-04 \\
        100 & 8000 & \SI{2303.86}{\second} & \SI{89.19}{\second} & \SI{28.48}{\second} & \SI{144.12}{\second} & \SI{13.14}{\second} & \SI{10.50}{\second} & 1.24e-04 & 2.31e-05 \\
        \hline
        \end{tabular}
    }
    \caption{Computational cost for GP training, prediction, and sampling. The standard approach using the Cholesky decomposition in single precision is compared to a fast implementation using batch GPs.
    We take $50$ gradient steps for training and predict/sample on 5000 test points.
    The mean and variance MAE between the two approaches is shown
    in the two rightmost columns.}
    \label{tab:batchGPs}
\end{table*}

\section{Details on \ALG}
\label{sec:scbo_details}
In all experiments we use the following hyperparameters for \ALG that were adopted from \TURBO \citep{eriksson2019}: $\tau_s = \max(3, \lceil d/10 \rceil)$, $\tau_f = \lceil d / q \rceil$, $L_{\min} = 2^{-7}$, $L_{\text{max}} = 1.6$, $\len_{\textrm{init}} = 0.8$, and perturbation probability $p_{\text{perturb}} = \min\{1, 20/d\}$, where $d$ is the number of dimensions and $q$ is the batch size.
Note that $\tau_s = 3$ is used by \citep{eriksson2019}, while we find that \ALG achieves better performance if we expand the trust region less frequently.
Recall that we assume the domain has been scaled to the unit hypercube $[0, 1]^d$.
A \emph{success} occurs if at least one evaluation in the batch improves on the incumbent.
In this case, we increment the success counter and reset the failure counter to zero.
If no point in the batch improves the current best solution, we set the success counter to zero and increment the failure counter.
We increase the side length to $\min(2L, L_{\text{max}})$ and reset botch counters to zero if the success counter reaches $\tau_s$.
Similarly, if the failure counter reaches $\tau_f$, we set the side length to $L/2$ and reset both counters.
If $L < L_{\min}$, we terminate the trust region and spawn a new trust region that is initialized with a Sobol sequence.
This trust region does not use any data collected by previous trust regions.
The discretized candidate set of size $r$ is also generated following the approach of \citet{eriksson2019} and we use $r=\min(200d, 5000)$ for all experiments.
In particular, we first create a scrambled Sobol sequence within the intersection of the trust region and the domain $[0, 1]^d$.
We use the value in the Sobol sequence with probability $p_{\text{perturb}}$ for a given candidate and dimension, and the value of the center otherwise.

\section{Global Consistency of \ALG}
\label{sec:global_convergence}
We prove that \ALG converges to a global optimum as the number of samples tends to infinity.
\begin{theorem}
    Suppose that \ALG with default parameters is used in a multi-start framework under the following conditions:
    \begin{enumerate}
        \item The initial points $\{y_i\}$ for \ALG are chosen such that for any $\delta > 0$ and $x \in [0,1]^d$ there exists $\nu(x, \delta) > 0$ such that the probability that at least one point in $\{y_i\}$ ends up in a ball centered at $x$ with radius $\delta$ is at least $\nu(x, \delta)$.

        \item The objective and constraints are bounded.

        \item There is a unique global minimizer $x^*$.

        \item \ALG considers any sampled point an improvement \emph{only if} it improves the current best solution by at least some constant $\gamma > 0$.
    \end{enumerate}
    Then, \ALG with noise-free observations converges to the global minimizer $x^*$.
\end{theorem}
Note that condition~(1) is met if the initial set is chosen uniformly at random.
Conditions (2) and (3) hold almost surely under the prior for our domain and are   common assumptions in global optimization \citep[e.g., see][]{regis2007stochastic,spall2005introduction}.
Condition~(4) is a straightforward design decision of the algorithm.

\begin{proof}
First observe that \ALG will take only a finite number of samples for any trust region due to conditions (2) and~(4).
Thus, \ALG will restart infinitely often with a fresh trust region and hence there is an infinite subsequence $\{x_k(i)\}$ of initial points.
This subsequence satisfies condition~(1) by design.
Thus, global convergence follows from the proof of global convergence for random search under condition~(3) \citep[e.g., see][]{spall2005introduction}.
\end{proof}
Note that this result also implies consistency for the~\TURBO algorithm of~\citet{eriksson2019}.

\section{Additional ablation studies}
\label{section_additional_ablation}
\begin{figure}[!ht]
    \centering
    \includegraphics[width=0.48\textwidth]{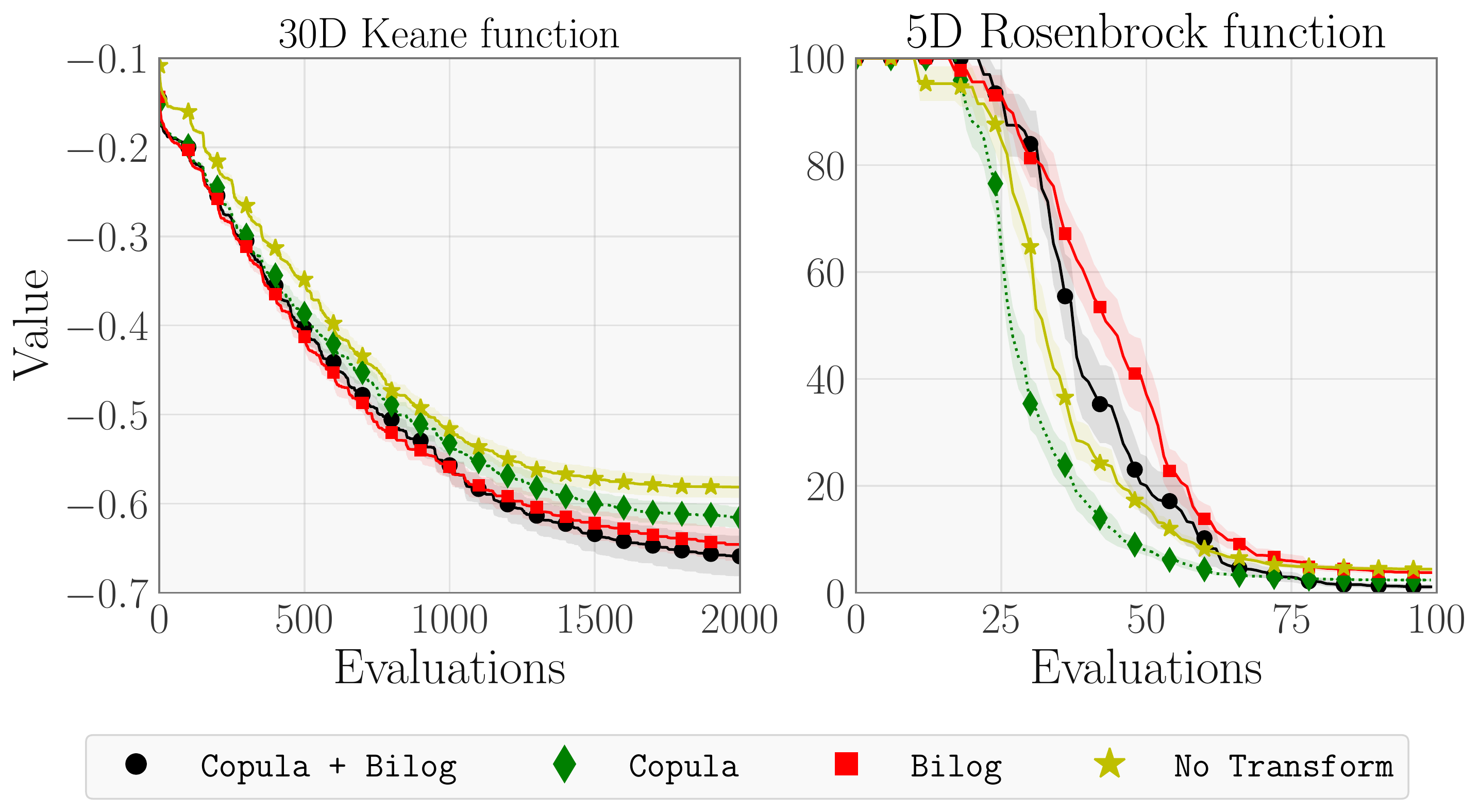}
    \caption{Individual contribution of the copula and bilog transforms in \ALG.
        \textbf{(Left)} The bilog transform is more important on the $30D$ Keane function.
        \textbf{(Right)} The copula transform improves the results on the $5D$ Rosenbrock function.
    }
    \label{fig:transforms}
    \raggedbottom
\end{figure}

In this section we provide two additional ablation studies that investigate the components of \ALG.
In Fig.~\ref{fig:transforms} we look at the effect of the copula and bilog transforms that are used by \ALG.
We note that the bilog transform is more important on the $30$D Keane function and that using both transforms works better than only using one of them.
The Copula transform improves the performance on the Rosenbrock problem.
Even though using the bilog transformation slows down the convergence, the final performance achieved by \ALG (that uses both transformations) is comparable to only using the Copula transformation.

\begin{figure}[!ht]
    \centering
    \includegraphics[width=0.38\textwidth]{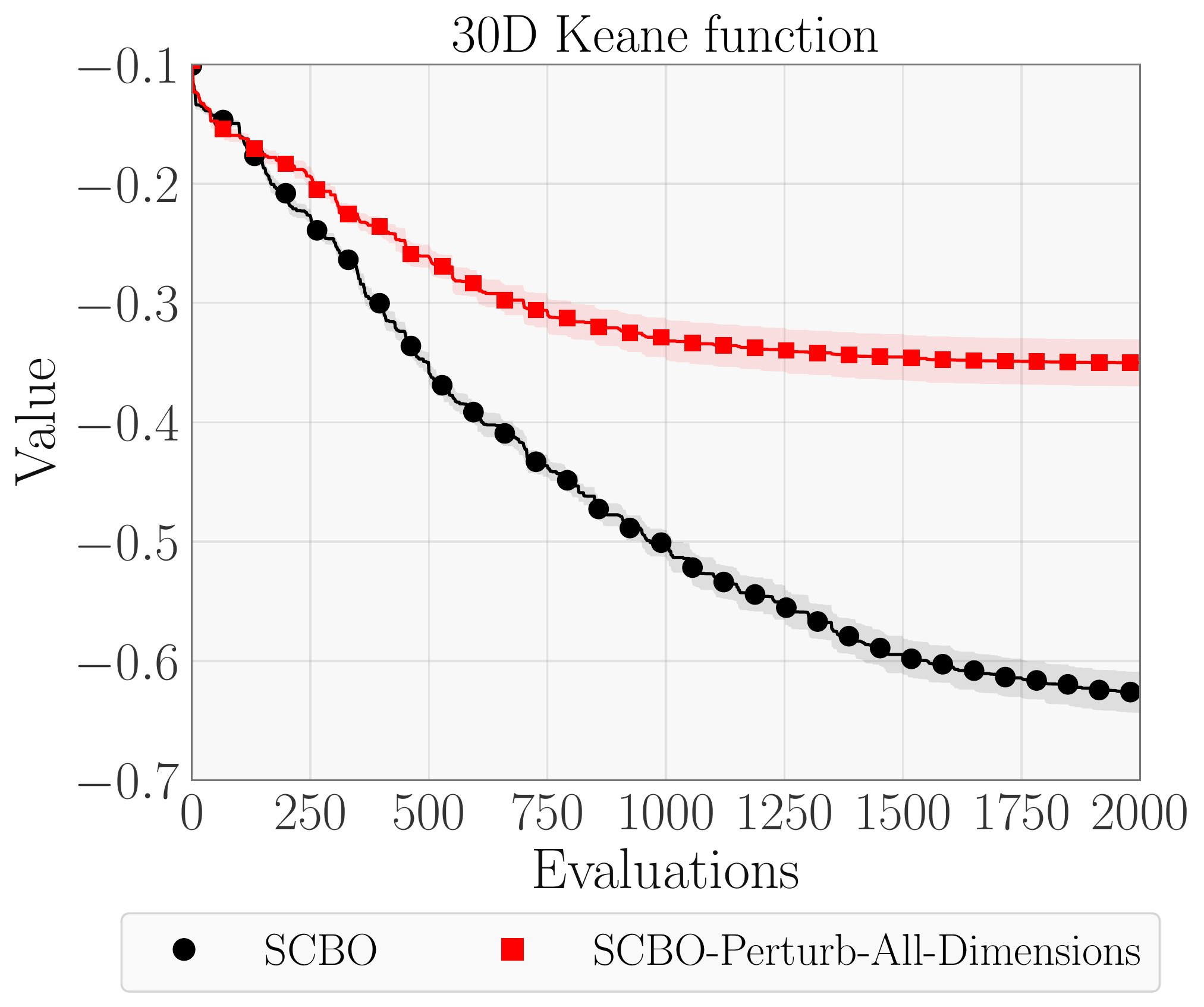}
    \caption{Results on the $30$D Keane function that illustrates the importance of only perturbing a subset of dimensions when generating the candidate set.
    }
    \label{fig:keane_perturb}
    \raggedbottom
\end{figure}

Another important component of \ALG is the perturbation probability from Sect.~\ref{sec:scbo_details}.
Perturbing only a subset of the dimensions helps \ALG generate candidates that are closer to the current best point which improves the performance on the $30$D Keane function, as we see in Fig.~\ref{fig:keane_perturb}.

\section{Results on the Transformed Physics Problems}
\label{sec:physics_transformed}
\begin{figure*}[!ht]
    \centering
    \includegraphics[width=0.80\textwidth]{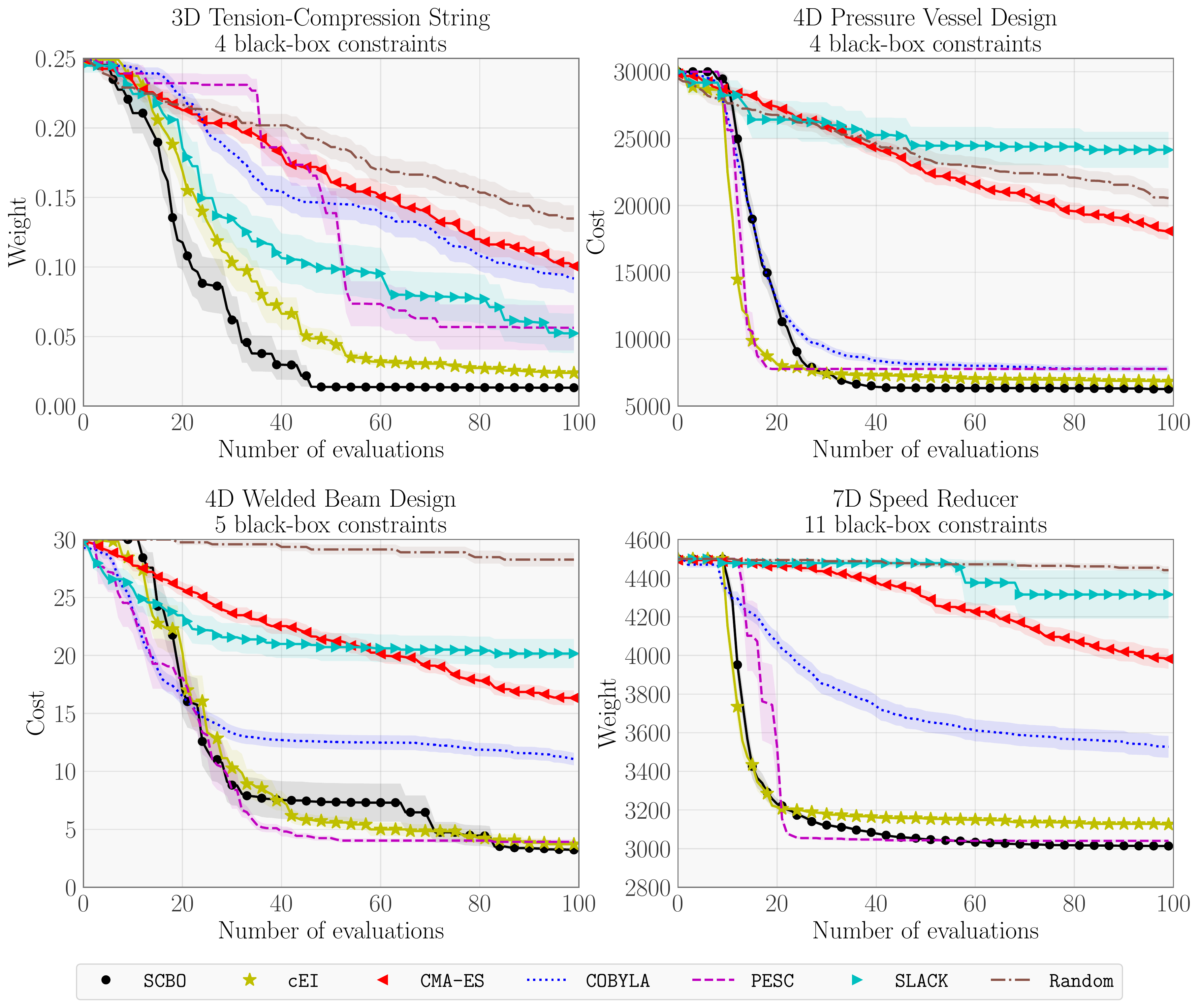}
    \caption{Results on the transformed physics problems, where the transformations of \ALG have been applied directly to the objective functions and constraints.
        \textbf{(Upper left)} \ALG and \CEI outperform the other methods on the Tension-compression string problem.
        \textbf{(Upper right)} \ALG finds the best solutions on the pressure vehicle design problem, followed by \CEI, \PESC, and \COBYLA.
        \textbf{(Lower left)} \PESC, and \CEI are eventually outperformed by \ALG on the welded beam design problem.
        \textbf{(Lower right)} \ALG and \PESC perform the best on the speed reducer problem.
    }
    \label{fig:physics_transformed}
    \raggedbottom
\end{figure*}

Recall that \ALG applies the \texttt{copula} transformation to the objective function and the \texttt{bilog} transformation to each constraint.
In this section we study on the four physics benchmarks how the performance of all baselines is affected by these transformations.
Fig.~\ref{fig:physics_transformed} summarizes the performances.
We note that the transformations do not lead to noticeable changes in performance for the baselines, except for \PESC that benefits on the 3D tension-compression string problem and on the 4D welded beam design.
The performance was comparable with and without the transformations for the other test problems.

\label{section_transformed_physics_problems}

\section{Details on the Benchmarks}
\label{sec:test_problems}
In this section we provide additional information for the test problems.
We refer the reader to the original papers for more details.

\subsection{3D Tension-Compression String}
The tension-compression string problem was described by \citet{hedar2006derivative}.
The goal is to minimize the weight subject to constraints on minimum deflection, shear stress, surge frequency, limits on outside diameter, and on design variables \citep{coello2002constraint}.
The first constraint is very sensitive to changes in the input parameters and cannot be modeled accurately by a global GP model.

\subsection{4D Pressure Vessel Design}
This problem was studied by \citet{coello2002constraint} and has four constraints.
The original problem does not specify bound constraints, so we use $0 \leq x_1, x_2 \leq 10$, $10 \leq x_3 \leq 50$, and $150 \leq x_4 \leq 200$.
This domain contains the best solution found by \citet{coello2002constraint}.
The goal is to minimize the total cost of designing the vessel, which includes the cost of the material, forming, and welding.
The variables describe the thickness of the shell, thickness of the head, inner radius, and length of the cylindrical section of the vessel.
The thickness of the shell and thickness of the head have to be multiples of 0.0625 and are rounded to the closest such value before evaluating the objective and constraints.

\subsection{4D Welded Beam Design}
This problem was considered by \citet{hedar2006derivative} and has 5 constraints.
The objective is to minimize the cost subject to constraints on shear stress, bending stress in the beam, buckling load on the bar,
end deflection of the beam, and three additional side constraints.

\subsection{7D Speed Reducer}
The 7D speed reducer model has 11 black-box constraints and was described by \citet{lemonge2010constrained}.
The objective is to minimize the weight of a speed reducer.
The design variables are the face width, the module of teeth, the number of
teeth on pinion, the length of shaft one between the bearings, the length of shaft two between the bearings, the diameter of shaft one, and the diameter of shaft two.

\subsection{10D Ackley}
In this problem we consider the popular 10-dimensional Ackley function
\begin{align*}
    f(x) = -20&\exp\left(-0.2\sqrt{\frac{1}{d} \sum_{i=1}^d x_i^2}\right) -\\
            &\exp\left(\frac{1}{d} \sum_{i=1}^d \cos(2\pi x_i)\right) + 20 + \exp(1)
\end{align*}
in the domain $[-5, 10]^{10}$ subject to the constraints $c_1(x) = \sum_{i=1}^{10} x_i \leq 0$ and $c_2(x) = \|x\|_2 - 5 \leq 0$.
This function is multi-modal and hard to optimize.
Additionally, the size of the feasible region is small making this problem even more challenging.
The optimal value is zero and is attained at the origin.

\subsection{30D Keane Bump}
For the Keane bump benchmark \citep{keane1994experiences},
the goal is to minimize
\[
    f(x) = -\left|\frac{\sum_{i=1}^d \cos^4(x_i) - 2 \prod_{i=1}^d \cos^2(x_i)}{\sqrt{\sum_{i=1}^d i x_i^2}}\right|
\]
subject to~$c_1(x) = 0.75 - \prod_{i=1}^d x_i \leq 0$ and~$c_2(x) = \sum_{i=1}^d x_i - 7.5d \leq 0$
over the domain $[0, 10]^d$.
We consider the case $d=30$ in our experiment.
The Keane benchmark is notoriously famous for being challenging for Bayesian optimization as it is hard to model with a global GP model.

\subsection{12D Robust Multi-point Optimization}
The goal is to learn a robust controller for the lunar lander in the OpenAI gym\footnote{\url{https://gym.openai.com/envs/LunarLander-v2}}.
The state space for the lunar lander is the position, angle, time derivatives, and whether or not either leg is in contact with the ground.
For each frame there are four possible actions: firing a booster engine left, right, up, or doing nothing.
The original objective was to maximize the average final reward over $m$ randomly generated terrains, initial positions, and velocities.
We extend this formulation to be more robust by adding $m$ constraints that no individual reward is below 200, which asserts that the lunar lander successfully lands in every scenario.
Moreover, we fix the~$m$ terrains throughout the optimization process therefore making the function evaluations deterministic (that is, without noise).

\subsection{60D Rover Trajectory Planning}
This problem was considered by~\citet{wang2017batched}.
The goal is to optimize the trajectory of a rover, where the trajectory is determined by fitting a B-spline to 30 design points in a 2D plane.
The reward function is $f(\vec x) = c(\vec x) + 5$,
where $c(\vec x)$ penalizes any collision with an object along the trajectory by ${-20}$.
We add the constraint that the trajectory has to start and end at the pre-specified start and end locations.
Additionally, we add 15 additional obstacles that are impassable and introduce constraints $c_i(x)$ for each $i$-th obstacle $o_i$ based on the final trajectory $\gamma(x)$ as follows:
\[
    c_i(x) =
    \begin{cases}
        -d(o_i, \gamma(x))  \quad \quad \quad \quad \quad \quad \text{ if } \gamma(x) \cap o_i = \emptyset, \\
        \displaystyle\max_{\alpha \in \gamma(x) \cap o_i} \displaystyle\min_{\beta \in \partial o_i} d(\alpha, \beta)  \,\,\,\,\, \quad \text{ otherwise},
    \end{cases}
\]
where $\partial o_i$ is the boundary of $o_i$.
That is, trajectories that do not collide with the object will be feasible under this constraint with a constraint value equal to the minimum distance between the trajectory and the object.
Trajectories that collide with the object will be given a larger constraint value if they intersect close to the center of the object.

\subsection{124D Vehicle Design with 68 Constraints (MOPTA08)}
MOPTA08 is a large-scale multi-disciplinary optimization problem from the vehicle industry~\citep{mopta08}.
There are~$124$ variables that describe gages, materials, and shapes as well as~$68$ performance constraints.
Note that a problem with this many input dimensions and black-box constraints is out of reach for existing methods in Bayesian optimization.

\subsection{2D Toy problem}
This problem was proposed by \citet{pesc}.
The goal is to minimize the function $f(x) = x_1 + x_2$ subject to \mbox{$c_1(x) = 1.5 - x_1 - 2x_2 - 0.5 \sin(2\pi(x_1^2 - 2x_2)) \leq 0$} and $c_2(x) = x_1^2 + x_2^2 - 1.5 \leq 0$.
The objective and constraints are all smooth low-dimensional functions.
The domain is the unit square $[0, 1]^2$.

\subsection{5D Rosenbrock function}
The goal is to minimize the Rosenbrock function \\ $f(x) = \sum_{i=1}^4 \left[100(x_{i+1} - x_i^2)^2 + (x_i - 1)^2 \right]$ subject to two constraints involving the Dixon-Price\footnote{\url{https://www.sfu.ca/~ssurjano/dixonpr.html}} (DP) function $c_1(x) = f_{\text{DP}}(x) - 10 \leq 0$ and the Levy\footnote{\url{https://www.sfu.ca/~ssurjano/levy.html}} function $c_2(x) = f_{\text{Levy}}(x) - 10 \leq 0$.
We created this problem to illustrate a setting where the objective and constraints are poorly scaled.
\ALG excels on this problem as the use of the trust region and robust transformations makes it possible to quickly make progress.
The domain is $[-3, 5]^5$.

\subsection{Restart Frequency of \ALG}
The \ALG algorithm \emph{restarts} with a new trust region if it fails to sample a better point for a certain number of consecutive steps; see Sect.~\ref{sec:scbo_details} and Sect.~3 of the main document for more details.
The average number of iterations between restarts with a new trust region is $42$ for the $2$D Toy problem, $88$ for the $5$D Rosenbrock function, $2095$ for the $30$D Keane function, and $2710$ for the $60$D Rover problem.